\documentclass[10pt,twocolumn,letterpaper]{article}

\usepackage{iccv}
\usepackage{times}
\usepackage{epsfig}
\usepackage{graphicx}
\usepackage{amsmath}
\usepackage{amssymb}

\usepackage{booktabs}
\usepackage{multirow}
\usepackage{float}
\usepackage{subfig}

\usepackage[pagebackref=true,breaklinks=true,letterpaper=true,colorlinks,bookmarks=false]{hyperref}

 \iccvfinalcopy 

\ificcvfinal\fi

\usepackage{color}
\definecolor{applegreen}{rgb}{0.55, 0.71, 0.0}

\begin{document}

\title{LPFF: A Portrait Dataset for Face Generators Across Large Poses}

\author{Yiqian Wu$^{1,2}$\qquad Jing Zhang$^{1,2}$\qquad Hongbo Fu$^{3}$\qquad Xiaogang Jin$^{1,2}$\thanks{Corresponding author.} \\
$^1$State Key Lab of CAD\&CG, Zhejiang University \\
$^2$ZJU-Tencent Game and Intelligent Graphics Innovation Technology Joint Lab\\
$^3$City University of Hong Kong \\
{\tt\small onethousand@zju.edu.cn,jing\_z99@163.com,hongbofu@cityu.edu.hk,jin@cad.zju.edu.cn}
}




\maketitle
\ificcvfinal\thispagestyle{empty}\fi

\begin{abstract}
The creation of 2D realistic facial images and 3D face shapes using generative networks has been a hot topic in recent years. Existing face generators exhibit exceptional performance on faces in small to medium poses (with respect to frontal faces) but struggle to produce realistic results for large poses. The distorted rendering results on large poses in 3D-aware generators further show that the generated 3D face shapes are far from the distribution of 3D faces in reality. We find that the above issues are caused by the training dataset's pose imbalance.

In this paper, we present \textit{LPFF}, a large-pose Flickr face dataset comprised of 19,590 high-quality real large-pose portrait images. We utilize our dataset to train a 2D face generator that can process large-pose face images, as well as a 3D-aware generator that can generate realistic human face geometry. To better validate our pose-conditional 3D-aware generators, we develop a new FID measure to evaluate the 3D-level performance. Through this novel FID measure and other experiments, we show that \textit{LPFF} can help 2D face generators extend their latent space and better manipulate the large-pose data, and help 3D-aware face generators achieve better view consistency and more realistic 3D reconstruction results.
\end{abstract}

\begin{figure}[h]
    	\centering
    	{\includegraphics[width=0.95\columnwidth]{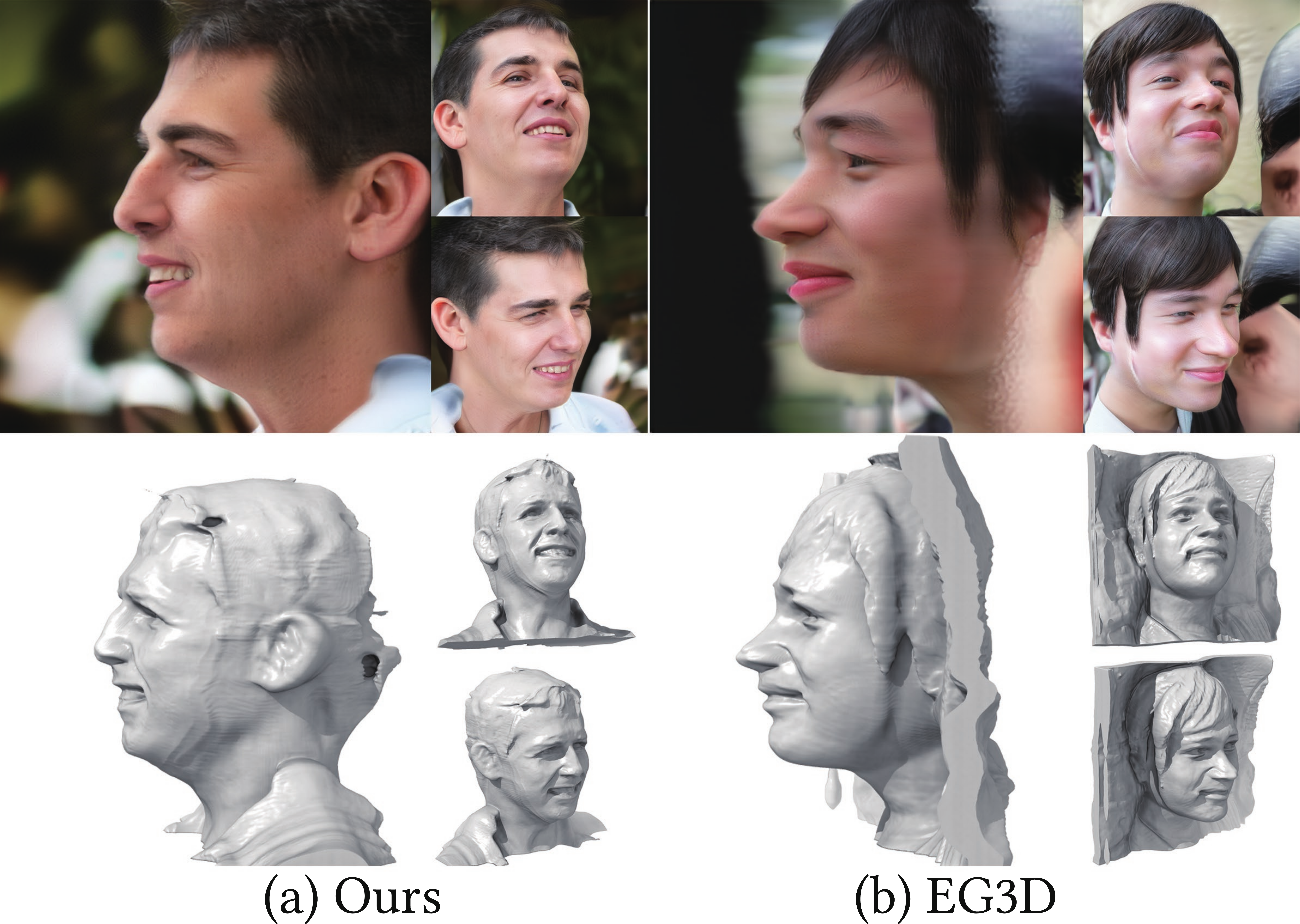}}
    	\caption{ Image and shape samples generated by EG3D models \cite{Chan_2022_CVPR} trained with the same training strategy but using different datasets (our new dataset \textit{LPFF} and \textit{FFHQ} for (a) and \textit{FFHQ} for (b)). 
      The generators are conditioned by the average camera parameters. 
      Shapes are iso-surfaces extracted from the corresponding density fields using marching cubes. Our dataset helps reduce distorted, ``seam'', ``wall-mounted'', and blurry artifacts exhibited in (b).}
      \vspace{-5pt}
    	\label{fig:teaser}
    \end{figure}

\section{Introduction}
    \label{sec:intro}
    Since the first introduction by Goodfellow in 2014,
    generative adversarial networks (GANs) \cite{DBLP:conf/nips/GoodfellowPMXWOCB14} have significantly advanced the performance of 2D high-resolution image generation. 
    GANs can accomplish a variety of downstream image editing tasks, particularly face modification \cite{DBLP:journals/tog/AbdalZMW21,nerffaceediting,DBLP:journals/corr/abs-2205-15517,DBLP:conf/cvpr/SunWZLZLW22}, thanks to the excellent image quality and semantic features in its latent space. 
    Recently, plenty of 3D-aware generators \cite{DBLP:conf/iclr/GuL0T22,DBLP:journals/corr/abs-2112-11427,Chan_2022_CVPR,DBLP:journals/corr/abs-2110-09788,DBLP:journals/corr/abs-2206-07695,DBLP:conf/cvpr/DengYX022,DBLP:journals/corr/abs-2206-10535,DBLP:journals/corr/abs-2301-09091} have been proposed to learn 3D-consistent face portrait generation from 2D image datasets. 
    3D-aware generators can describe and represent geometry in their latent space while rendering objects from different camera perspectives using volumetric rendering. Researchers carefully designed generator architectures and training strategies to accelerate training, reduce memory overheads, and increase rendering resolution.
     
\begin{figure}[t]
          \centering
          \includegraphics[width=.85\columnwidth]{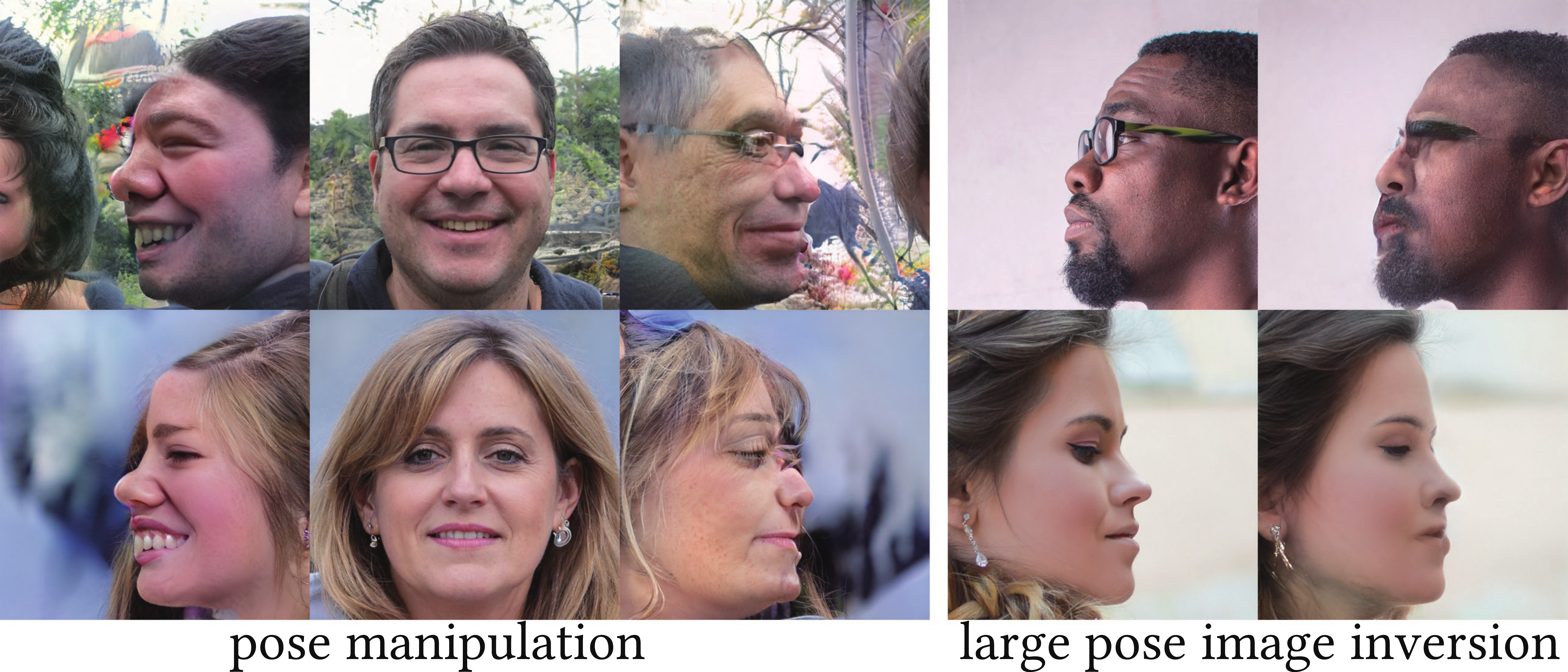}
          \caption{
          StyleGAN2 \cite{DBLP:conf/cvpr/KarrasLAHLA20}'s large-pose performance when trained on \textit{FFHQ}. InterfaceGAN \cite{DBLP:conf/cvpr/ShenGTZ20} is used to edit the yaw angle of randomly sampled latent codes. 
          {We use optimization-based GAN inversion to obtain the latent codes of target large-pose real images.}
          }
          \vspace{-5pt}
          \label{fig:2d_gans_steep_yaw-1}
        \end{figure}

\begin{figure}[t] 
          \centering
          \includegraphics[width=.85\columnwidth]{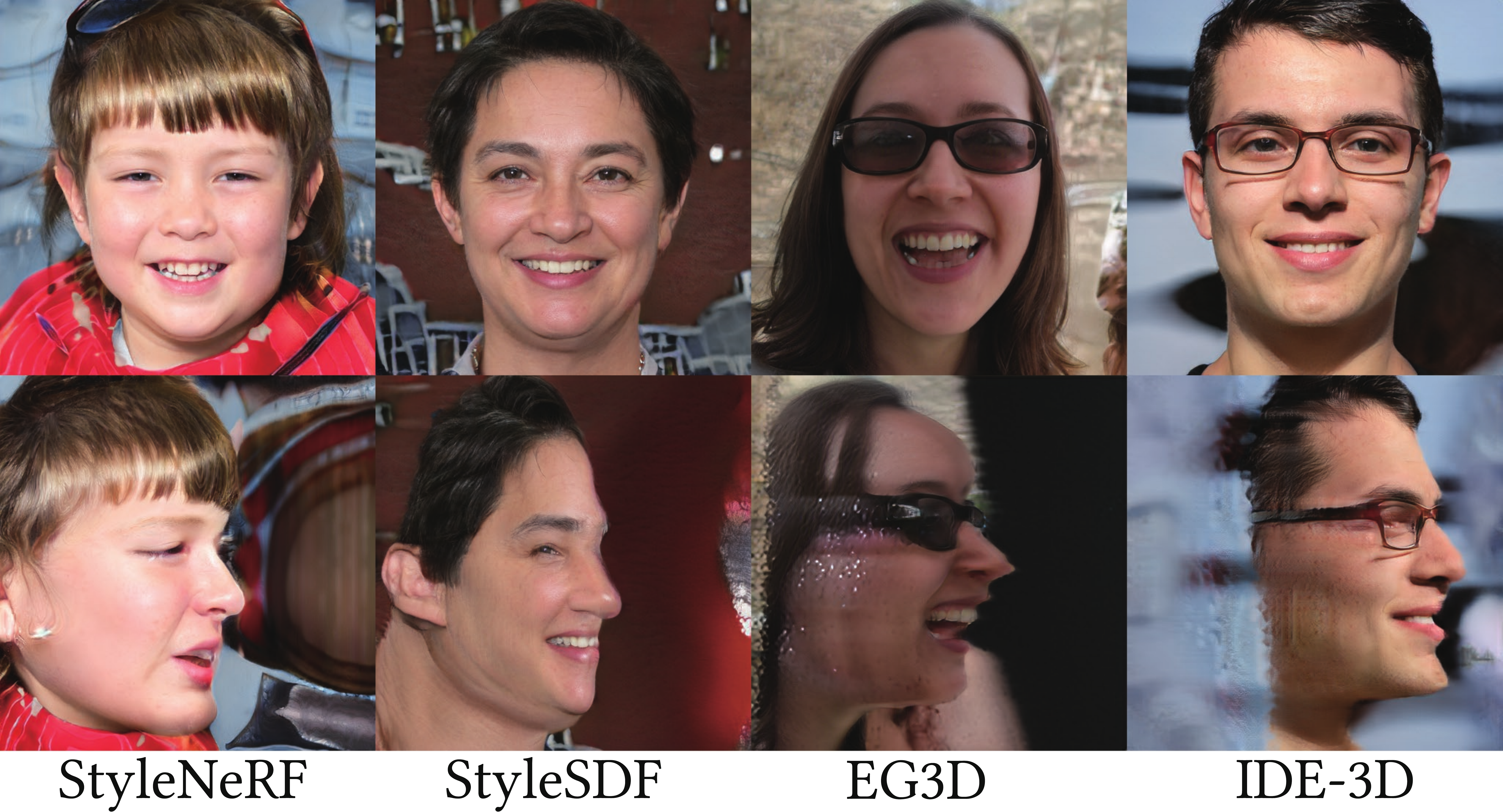}
          \caption{
          3D-aware generators trained on \textit{FFHQ} (StyleNeRF \cite{DBLP:conf/iclr/GuL0T22}, StyleSDF \cite{DBLP:journals/corr/abs-2112-11427}, EG3D \cite{Chan_2022_CVPR}, and IDE-3D \cite{DBLP:journals/corr/abs-2205-15517}) achieve excellent image synthesis performance on faces in small to medium 
           poses (Top),  but exhibit obvious artifacts at steep angles (Bottom).} 
          \vspace{-5pt}
          \label{fig:ed_aware_gans_steep_yaw-1}
        \end{figure}

    Both the existing 2D and 3D approaches, however, are unable to process large-pose face data.
    Regarding 2D face generators, those large-pose data are actually outside of their latent space, which prevents them from generating reasonable large-pose data, thus causing at least two problems. 
    First, as shown in Fig.~\ref{fig:2d_gans_steep_yaw-1} (left), moving the latent code along the yaw pose editing direction will cause it to reach the edge of the latent space before faces become profile.
    Second, as shown by the results of image inversion
    in Fig.~\ref{fig:2d_gans_steep_yaw-1} (right), it is challenging to project large-pose images to the latent space, let alone perform semantic modification on them.
    One of the goals of 3D-aware generators is to model realistic human face geometry, but existing 3D-aware generators trained on 2D image datasets still have difficulty producing realistic geometry. This issue is more serious when rendering the results at extreme poses.
    As shown in Fig.~\ref{fig:ed_aware_gans_steep_yaw-1}, faces synthesized by those methods have noticeable artifacts, including distortion, blurring, and stratification. 
    In Fig. \ref{fig:teaser} (b), EG3D shows a ``wall-mounted'' and distorted 3D representation without ears. All these indicate that the generated face shapes are not realistic enough.

    The above issues in the face generators are mainly caused by the unbalanced camera pose distribution of the {narrow-range} training dataset.
    \textbf{F}lickr-\textbf{F}aces-\textbf{HQ} Dataset (\textbf{FFHQ}) is a popular high-quality face dataset used to train those face generators, but it mainly contains images limited to small to medium poses.
    As a result, 2D and 3D-aware generators cannot learn a correct large-pose face
    distribution without sufficient large-pose data.
    To avoid artifacts under large poses, downstream applications \cite{DBLP:journals/corr/abs-2205-15517,DBLP:journals/corr/abs-2203-13441,DBLP:conf/wacv/KoCCRK23,xie2022high,nerffaceediting,DBLP:conf/siggrapha/JinRKBC22,DBLP:journals/corr/abs-2301-02700,DBLP:journals/corr/abs-2211-16927} based on those face generators typically sample small poses, which limits their application scenarios. 
    
    It is difficult to get a pose-balanced dataset. 
    First, large-pose faces are nearly impossible to detect using Dlib \cite{DBLP:conf/cvpr/KazemiS14}, a popular face detector, and the one used to crop \textit{FFHQ}.
    Second, simply replicating extremely limited large-pose data to balance the pose distribution is insufficient to help extend the camera distribution.
    As a result, it is critical to collect a large number of large-pose, {in-the-wild}, and high-resolution face images, which are lacking in existing datasets.
    
    In this paper, we propose a novel high-quality face dataset containing \textbf{19,590} real large-pose face images, named \textbf{L}arge-\textbf{P}ose-\textbf{F}lickr-\textbf{F}aces Dataset (\textbf{LPFF}), as a supplement to \textit{FFHQ}, in order to extend the camera pose distribution of \textit{FFHQ} and train 2D and 3D-aware face generators that are free of the aforementioned problems. 
    Given the difficulty of large-pose face detection and the imbalanced distribution of camera poses in real-life photographs, we design a face detection and alignment pipeline that is better suited to large-pose images.
    Our method can also gather large amounts of large-pose data based on pose density. We retrain StyleGAN2-ada \cite{DBLP:conf/nips/KarrasAHLLA20} to demonstrate how our dataset can assist 2D face generators in generating and editing large-pose faces. We retrain EG3D \cite{Chan_2022_CVPR} as an example to demonstrate how our dataset can aid 3D face generators in understanding realistic face geometry and appearance across a wide range of camera poses. 
    {
    In order to better evaluate the 3D-level performance of EG3D models trained on different datasets, we propose a new FID measure for pose-conditional 3D-aware generators.
    }
    Extensive experiments show that our dataset leads to realistic large-pose face generation and manipulation in the 2D generator. Furthermore, our dataset results in more realistic face geometry generation in the 3D-aware generator.

    Our paper makes the following major contributions: 
     1) {A novel data processing and filtering method that can collect large-pose face data from the Flickr website according to camera pose distribution, leading to a novel face dataset that contains 19,590 high-quality real large-pose face images.}
     2) A retrained 2D face generator that can process large-pose face images.
     3) A retrained 3D-aware generator that can generate realistic human face geometry.
     4) A new FID measure for pose-conditional 3D-aware generators.

   \begin{figure*}[t]
    	\centering
    	\subfloat[]{\includegraphics[width=.20\linewidth]{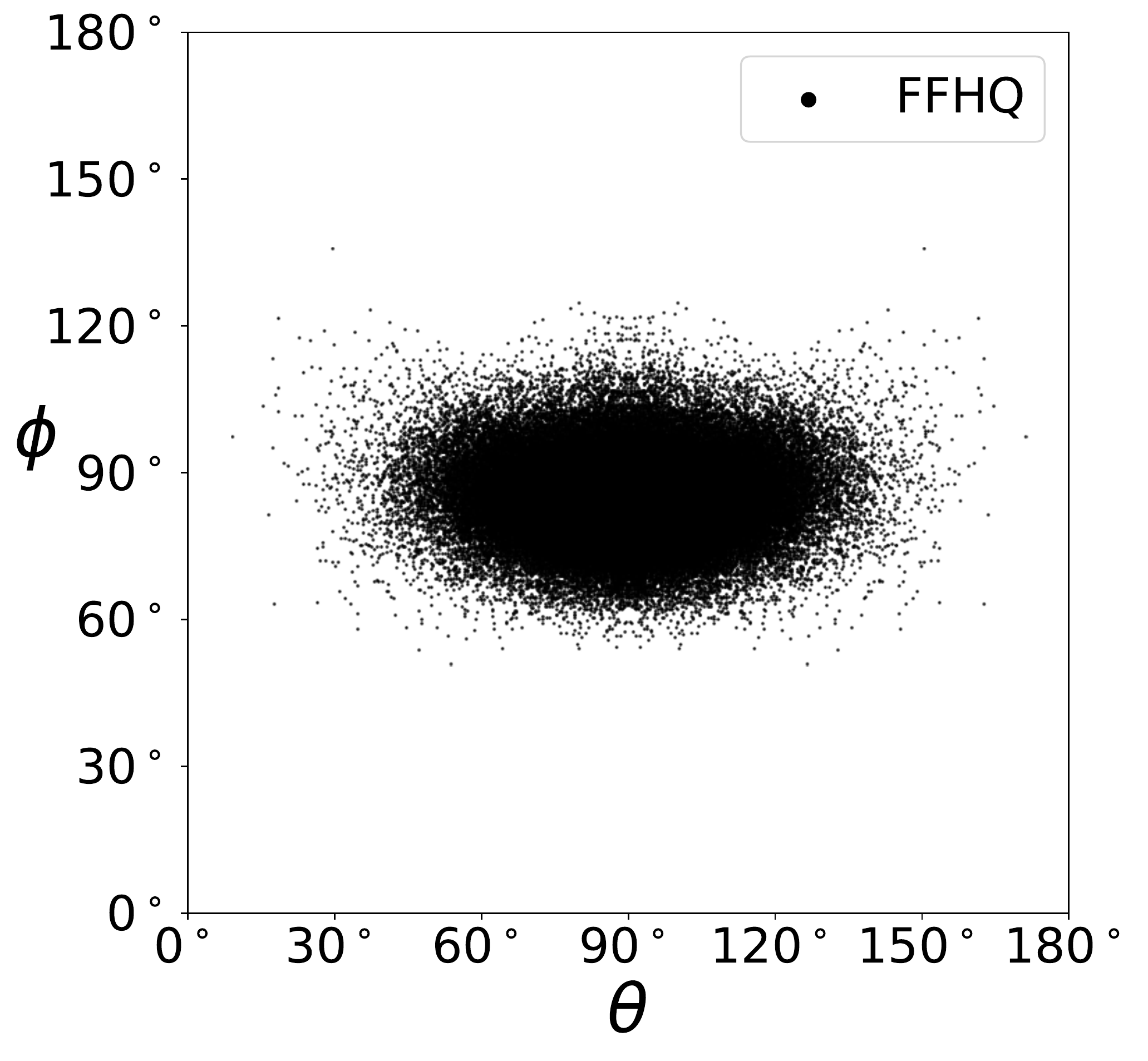}}\hspace{1pt}
    	\subfloat[]{\includegraphics[width=.20\linewidth]{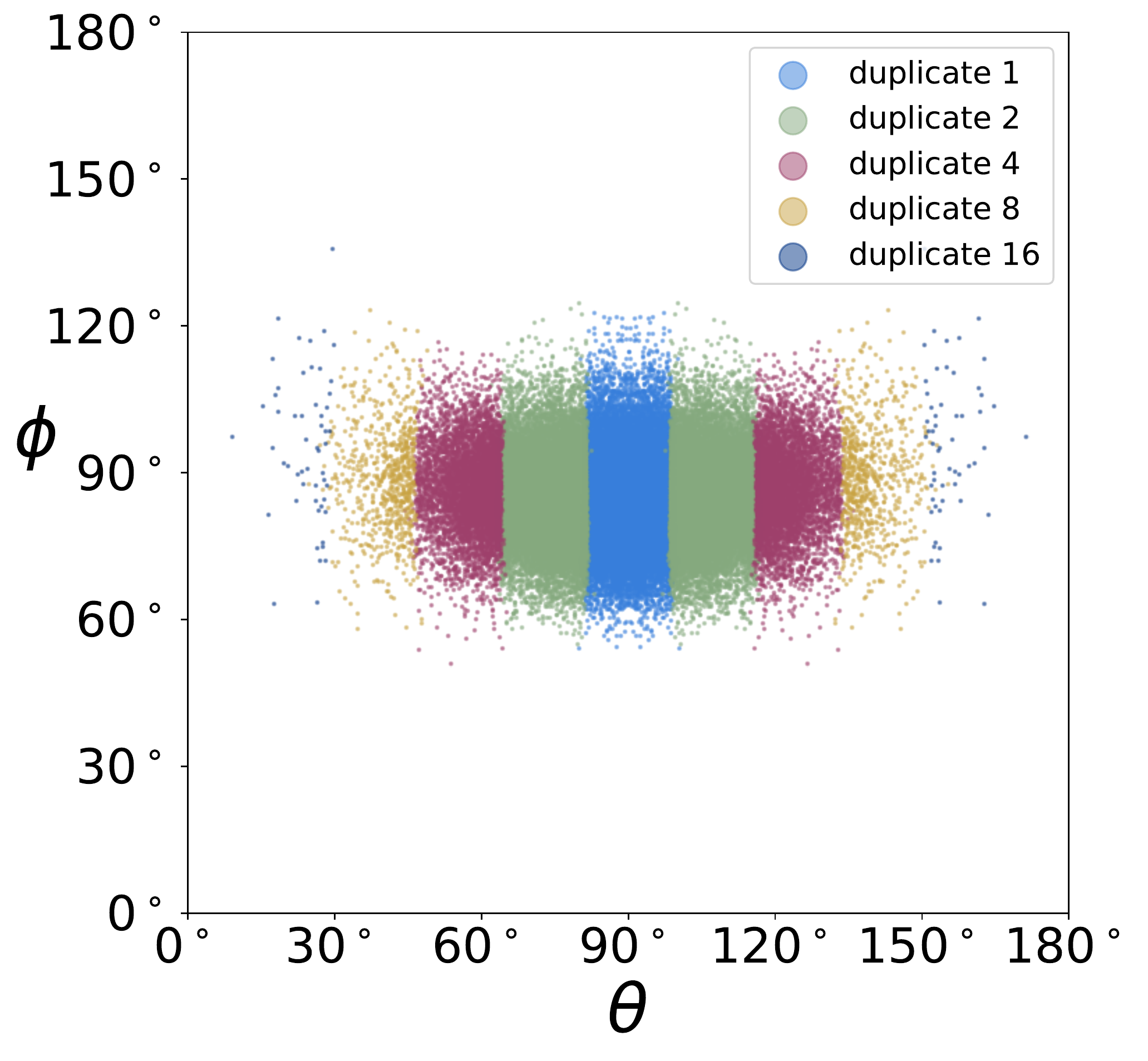}}\hspace{3pt}
    	\subfloat[]{\includegraphics[width=.20\linewidth]{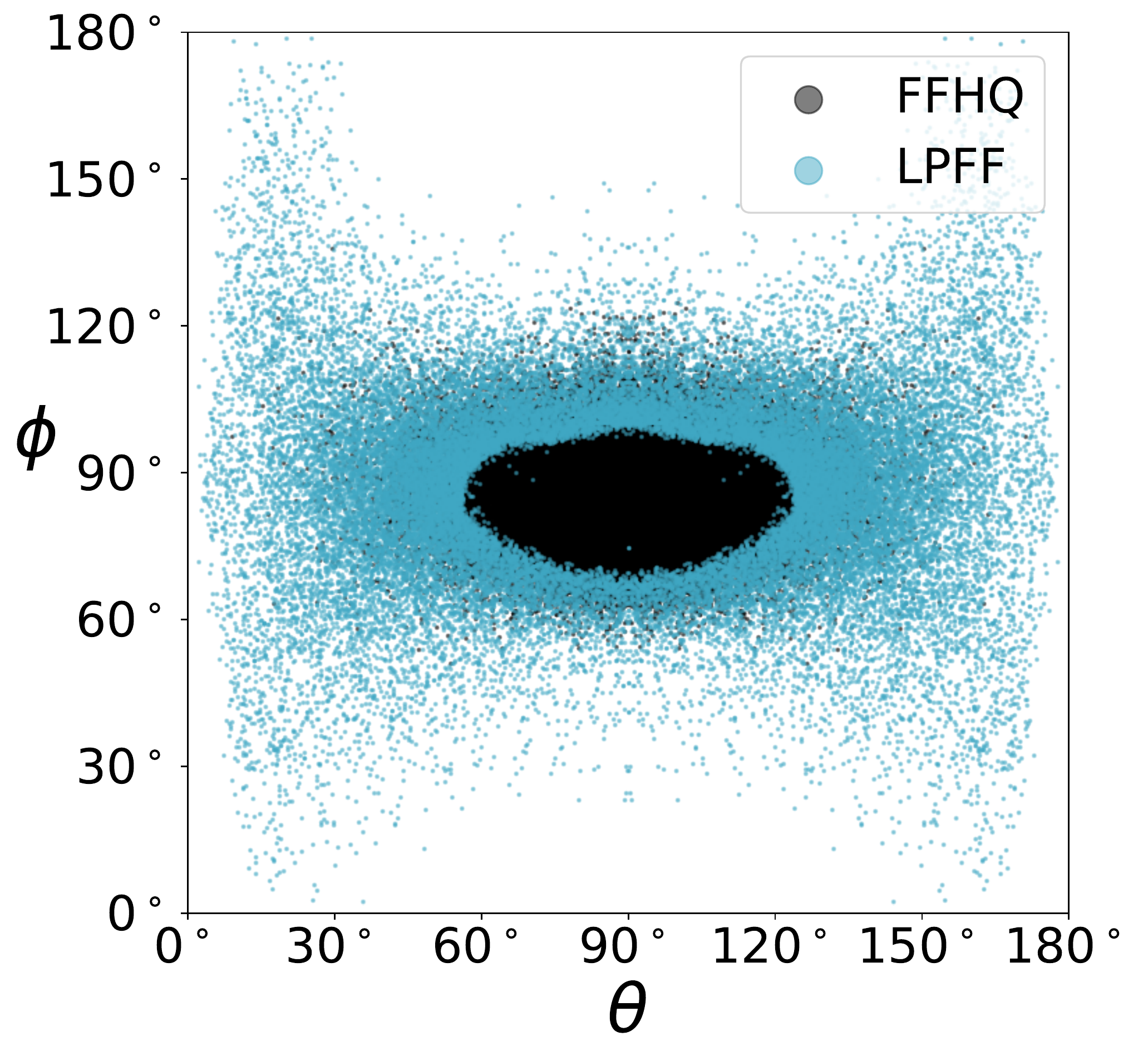}}\hspace{3pt}
    \subfloat[]{\includegraphics[width=.20\linewidth]{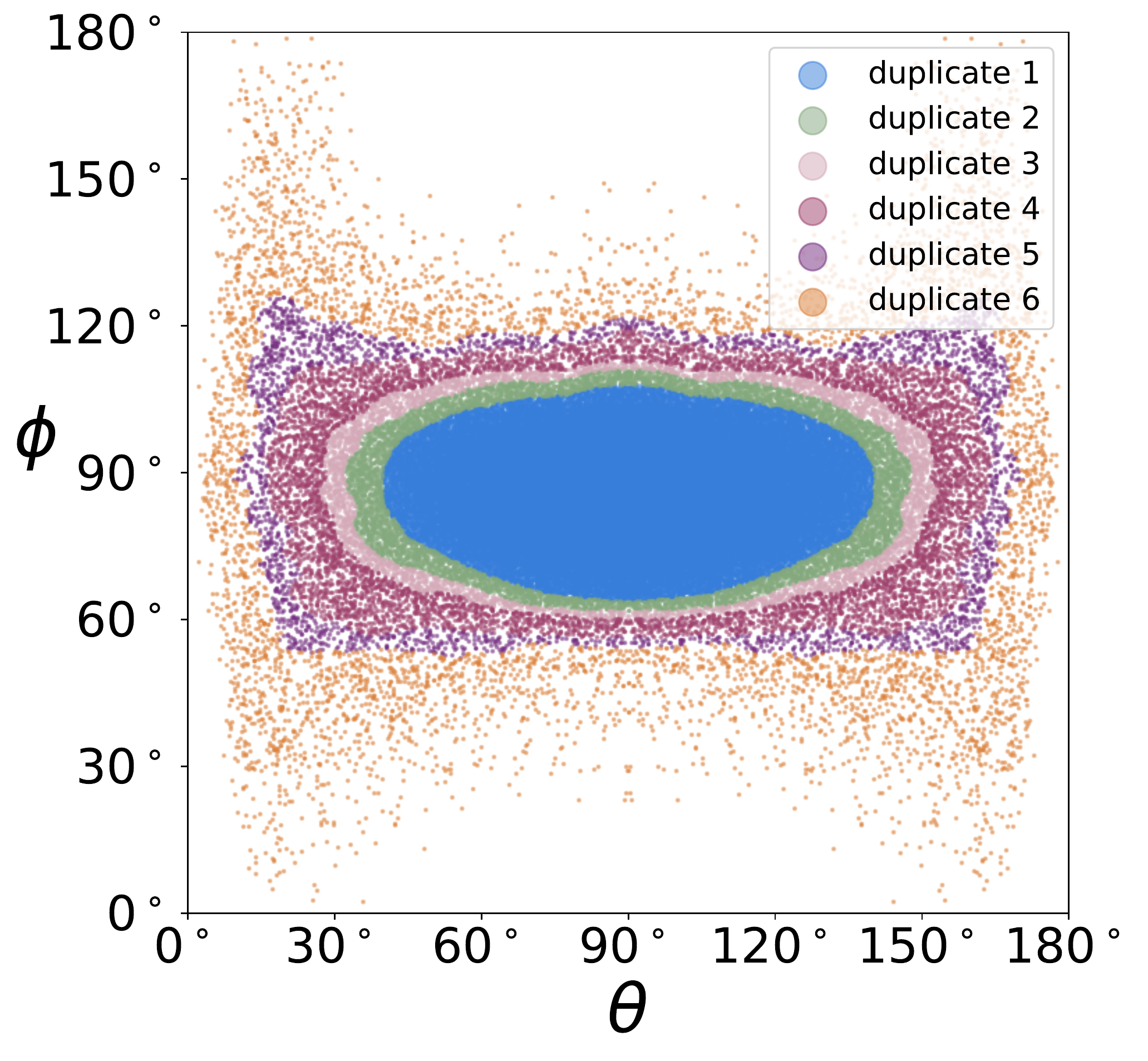}}
    
    	\caption{
    	(a) \textit{FFHQ}. (b) \textit{FFHQ-rebal}. (c) \textit{FFHQ+LPFF}. (d) \textit{FFHQ+LPFF-rebal}.
     {The term ``duplicate'' refers to the number of repetitions in data resampling.} }
     \vspace{-5pt}
    	\label{fig:data_distribution}
    \end{figure*}
    
\section{Related Work}
\label{sec:related_work}
    \paragraph{2D Face Generators.}
    Since Goodfellow first proposed the generative adversarial networks (GANs) in 2014  \cite{DBLP:conf/nips/GoodfellowPMXWOCB14}, a growing number of distinct GANs model designs \cite{DBLP:journals/corr/RadfordMC15,DBLP:conf/nips/GulrajaniAADC17,DBLP:conf/iclr/BrockDS19,DBLP:conf/iclr/KarrasALL18} have been developed to
    produce more impressive performance on realistic image synthesis. 
    Among these GANs models, StyleGAN \cite{DBLP:conf/cvpr/KarrasLA19,DBLP:conf/cvpr/KarrasLAHLA20,DBLP:conf/nips/KarrasAHLLA20,DBLP:conf/nips/KarrasALHHLA21} is regarded as the most cutting-edge generator of high-quality images. For face portrait images, StyleGAN provides not only realistic image generation but also implicit semantic features in latent space, which are
    beneficial for many downstream computer vision applications \cite{DBLP:journals/tog/AbdalZMW21,DBLP:journals/tog/ChenLLRLF021,DBLP:conf/iccv/PatashnikWSCL21}.
    However, the face StyleGAN is trained on a pose-imbalanced face dataset, \textit{FFHQ}. StyleGAN inherits the pose bias from \textit{FFHQ}, resulting in artifacts and distortions when projecting and editing large-pose portraits. This issue is especially noticeable in downstream practical applications because in-the-wild photographs are not always forward-looking.
    
    \paragraph{3D-aware Face Generators.}
    2D generators have been expanded to support 3D multi-view rendering.    
    Early methods combined voxel rendering \cite{DBLP:conf/nips/Nguyen-PhuocRMY20,DBLP:conf/iccv/HenzlerM019,DBLP:conf/nips/ZhuZZ00TF18} or NeRF rendering \cite{DBLP:conf/nips/SchwarzLN020,DBLP:conf/iclr/GuL0T22,DBLP:conf/cvpr/ChanMK0W21} with generators to support view-consistent image synthesis. 
    However, those methods have a high cost of calculation, which limits the resolution of the output.
    Later, a number of studies suggested an additional super-resolution network to enhance image quality without adding too much computing load \cite{Chan_2022_CVPR,DBLP:conf/siggraph/TanFMOTPTTZ22,DBLP:journals/corr/abs-2112-11427,DBLP:conf/cvpr/XueLSL22}. 
    Researchers also suggested an effective optimization strategy \cite{DBLP:journals/corr/abs-2206-10535} to directly output high-resolution results without any super-resolution module.
    These techniques not only produce view-consistent results, but also learn, represent, and generate face geometry in a generative manner.
    {Several methods \cite{DBLP:journals/corr/abs-2205-15517,nerffaceediting} achieve semantic attribute editing and geometry-appearance disentanglement in view-consistent image synthesis through the integration of semantic masks into 3D-aware generators.}
    However, for human face training, they rely heavily on 2D image collections (\textit{FFHQ}). Because of the pose imbalance in the training dataset, incorrect facial geometry may arise in the final results, which negatively impacts the performance of downstream applications \cite{DBLP:journals/corr/abs-2205-15517,DBLP:journals/corr/abs-2203-13441,DBLP:conf/wacv/KoCCRK23,xie2022high,nerffaceediting,DBLP:conf/siggrapha/JinRKBC22,DBLP:journals/corr/abs-2301-02700,DBLP:journals/corr/abs-2211-16927}.
    Researchers also try to eliminate the dependency for 3D pose priors \cite{shi2023pof3d} or resample the dataset to increase the density of extremely limited large-pose data \cite{Chan_2022_CVPR}. However, both of them cannot address the root causes.

    \paragraph{Face Image Datasets.}
     Numerous studies have noted the pose imbalance in current face image datasets.
     300W-LP \cite{DBLP:conf/cvpr/ZhuLLSL16} is a dataset consisting 
     of 61,225 images across large poses, but all the images {are artificially synthesized by face profiling}.
     AFLW \cite{DBLP:conf/iccvw/KostingerWRB11} contains 21,080 face images with large-pose variations, LS3D-W \cite{DBLP:conf/iccv/BulatT17} contains $\sim$230,000 images from a combination of 300-W test set \cite{DBLP:conf/cvpr/SagonasTZP13}, 300-VW \cite{DBLP:conf/iccvw/ShenZCKTP15}, Menpo \cite{DBLP:conf/cvpr/ZafeiriouTCDS17}, and AFLW2000-3D \cite{DBLP:conf/cvpr/ZhuLLSL16}. But most images in AFLW and LS3D-W are at low resolution.
     {There are several 3D face datasets \cite{DBLP:conf/cvpr/Yang0WHSYC20,wuu2022multiface,DBLP:journals/corr/abs-1904-00168} that contain high-quality multi-view face images mainly {for 3D face reconstruction}}. However,
    {these datasets have limited variety
    \cite{DBLP:conf/cvpr/Yang0WHSYC20}, or are low-resolution \cite{DBLP:conf/iccvw/KostingerWRB11,DBLP:conf/iccv/BulatT17}, or are synthesized artificially \cite{DBLP:conf/cvpr/ZhuLLSL16}. In contrast, our dataset consists of high-resolution real images collected from in-the-wild photographs.}

\section{Data Preparation}
\label{sec:data_preparation}
In this section, we will introduce how to build our
large-pose face dataset. First, we describe the process for extracting data density from \textit{FFHQ} (Sec.~\ref{subsec:Camera Parameter}). Then, we introduce a novel data processing pipeline that can produce more reasonable realigned results (Sec.~\ref{subsec:Data Processing}). 
In order to filter large-pose face data from the Flickr images according to camera distribution, we propose to employ the pose density function to collect only large-pose data (Sec.~\ref{subsec:Large Pose Data Selection}). Finally, we introduce a novel rebalancing strategy (Sec.~\ref{subsec:Data Rebalance}). 

\subsection{Camera Distribution}
\label{subsec:Camera Parameter}
   {EG3D uses a face reconstruction model \cite{DBLP:conf/cvpr/DengYX0JT19}, denoted as $\mathcal{F}$ in this paper, to extract camera parameters.
   All cameras are assumed to be positioned on a spherical surface with a radius $r=2.7$, and the camera intrinsics are fixed.}
   In this paper, we only consider the camera location and ignore the roll angle of the camera to compute the camera distribution {(detailed in the supplementary file)}. 
   We convert the coordinates of each camera in \textit{FFHQ} from Cartesian coordinates to spherical coordinates and get their $\theta$ and $\phi$ (see Fig. \ref{fig:data_distribution} (a)). Notice that the face with $\theta = 90^\circ$ and $\phi = 90^\circ$ is frontal.

\subsection{Data Processing}
\label{subsec:Data Processing}
    {Given the difficulty of large-pose face detection and the imbalanced distribution of camera poses in real-life photographs, we propose a novel mechanism to collect, process, and filter large-pose data.}
    We first collect \textbf{155,720} raw portrait images from Flickr\footnote{\hyperref[]{https://www.flickr.com}} (with permission to copy, modify, distribute, and perform). 
    Then we remove all the raw images that already appeared in \textit{FFHQ}.

    Our pipeline is based on that of EG3D, and we respectively align each raw image according to the image align function in EG3D and StyleGAN.
    In EG3D, the authors first predict the 68 face landmarks of a raw image by Dlib, 
    and then get a realigned image by using the eyes and mouth positions to determine a square crop window for cropping and rotating the raw image.
    The realigned image is denoted as $X_{realigned}$  with the eyes at the horizontal level and the face at the center of the image. 
    Then MTCNN \cite{7553523} is used to get the positions of the eyes, the nose, and the corners of the mouth of $X_{realigned}$, and the 5 feature points are then fed into $\mathcal{F}$ to predict camera parameters. Finally, these positions are used to crop $X_{realigned}$, resulting in the final image.  

    In our pipeline, we first use Dlib to get 68 landmarks for each of the 155,720 raw portrait images, and for those images that resist face detection, we additionally apply face alignment \cite{DBLP:conf/iccv/BulatT17} (SFD face detector) to predict landmarks.  
    The face alignment detector achieves better performance on large-pose face detection than Dlib.
    Joining the two landmark predictors can help us detect as many large-pose faces as possible. Then the predicted landmarks are used to get the realigned image $X_{realigned}$. 
    {In this step, we get \textbf{506,262} $X_{realigned}$.}
    
    We find that the MTCNN sometimes cannot predict landmarks for large-pose faces. So instead of using MTCNN, we directly aggregate the 68 landmarks to get the 5 feature points of the eyes, mouth, and nose.

    After that, we use $\mathcal{F}$ to predict camera parameters. 
    Then we filter large-pose face data from \textbf{506,262} $X_{realigned}$ (detailed in Sec. \ref{subsec:Large Pose Data Selection}), getting \textbf{208,543}  large pose $X_{realigned}$.
    We automatically filter out low-resolution images and manually examine the rendering results of the reconstructed face models, removing any failed 3D reconstructions (which indicate incorrectly estimated camera parameters), as well as blurry or noisy images. 
    Finally, we get \textbf{19,590}
    high-quality large-pose face images with correctly estimated camera parameters. 

    When cropping the final image, we find that some of the 5 feature points (especially when there is a face with eyeglasses) are not accurate enough to crop $X_{realigned}$ properly, but after manual selection,
    the landmarks that $\mathcal{F}$ produces are more aligned with the input faces. 
    {So we use the landmarks of the reconstructed face to crop $X_{realigned}$ according to EG3D and StyleGAN functions and obtain final images. Please refer to the supplement file for an illustration of the image processing pipeline.}

\subsection{Large-Pose Data Selection}
    \label{subsec:Large Pose Data Selection}
    {
   To collect only images with ``low density'' (at large poses), we propose using the density function of FFHQ to filter large pose faces.
    Inspired by \cite{DBLP:journals/tog/LeimkuhlerD21}, 
    we estimate the density of the \textit{FFHQ} camera $(\theta, \phi)$ tuples using Gaussian kernel density estimation and  Scott’s rule \cite{DBLP:books/wi/Scott92} as a bandwidth selection strategy. 
   After obtaining $\rho_{ffhq}$, where $density = \rho_{ffhq} (\theta, \phi)$ is the density of the camera at $(\theta, \phi)$, 
   we use $\rho_{ffhq}$ to compute the density of \textbf{506,262} $X_{realigned}$, and filter the images with a density less than 0.4 ($density = \rho_{ffhq} (\theta, \phi)<$  0.4).
   }

    \subsection{Data Rebalance}
    \label{subsec:Data Rebalance} 

    After image processing, large pose filtering, and carefully manual selecting, we get \textbf{19,590} large-pose face images as our \textit{LPFF} dataset.
    We use  the \textit{LPFF} dataset as a supplement to \textit{FFHQ}. That is, we combine \textit{LPFF} with \textit{FFHQ}, named \textit{FFHQ+LPFF}. The datasets are augmented by a horizontal flip.
    In Fig.~\ref{fig:data_distribution}, we show the camera distribution for both \textit{FFHQ+LPFF} and \textit{FFHQ}. 

    To improve our models' performance on large-pose rendering quality and image inversion, we propose using a resampling strategy to further rebalance our \textit{FFHQ+LPFF} dataset (refer to Sec. \ref{sec:Evaluation} for evaluation).
    In EG3D, in order to increase the sampling probability of the low-density data, the authors rebalanced the \textit{FFHQ} dataset by splitting it into 9 uniform-sized bins across the yaw range and duplicating the images according to the bins (as shown in Fig.~\ref{fig:data_distribution} (b)). We denoted the rebalanced \textit{FFHQ} dataset as \textit{FFHQ-rebal}.

    Inspired by EG3D, we also rebalance \textit{FFHQ+LPFF} to help the model focus more on large-pose data.
    Instead of simply splitting the dataset according to yaw angles, we split \textit{FFHQ+LPFF} according to the data densities (Fig.~\ref{fig:data_distribution} (d)). Similar to Sec.~\ref{subsec:Camera Parameter}, we first compute the pose density function of \textit{FFHQ+LPFF} (denoted as $density = \rho_{ffhq+lpff} (\theta, \phi)$), 
    then duplicate our dataset as:
    \begin{equation}
    \centering
        \left\{
            \begin{array}{lr}
                N = \mathop{\min}(\mathop{\max}(\mathrm{round}(\frac{\alpha}{density}), 1), 4),density \geq 0.03&   \\
                N = 5,density \in [0.02,0.03)&  \\
               N = 6,density \in [0,0.02)&    
            \end{array}
        \right.
    \end{equation}
    where $\alpha$ is a hyper-parameter (we empirically set $\alpha=0.24$ in our experiments), and $N$ denotes the number of repetitions. The rebalanced \textit{FFHQ+LPFF} is denoted as \textit{FFHQ+LPFF-rebal}.


\section{Training Details}
\label{sec:training_details}
In this section, we will retrain 2D and 3D-aware face generators using our dataset.
Regarding the 2D generator (Sec.~\ref{subsec:StyleGAN_retrain}), we retrain StyleGAN2-ada using our dataset before fine-tuning the model using the rebalanced dataset.
As for the 3D-aware generator (Sec.~\ref{subsec:EG3D_retrain}), we first use our dataset to retrain EG3D, and then use the rebalanced dataset to fine-tune the model. 
In order to improve image synthesis performance during testing, we further fine-tune the model by setting the camera parameters input to the generator as the average camera.

\subsection{StyleGAN}   
\label{subsec:StyleGAN_retrain}
    \paragraph{Retrain.}
    In the StyleGAN training, we use the StyleGAN2-ada architecture as our baseline, and train it on \textit{FFHQ+LPFF} from scratch. 
    We use the training parameters 
    defined by \textit{stylegan2} config in StyleGAN2-ada.
     We denote the StyleGAN2-ada model 
     trained on \textit{FFHQ} as $S^{FFHQ}_{var1}$, and the model 
     trained on \textit{FFHQ+LPFF} as $S^{Ours}_{var1}$. 
    Our training time is $\sim$5 days on 8 Tesla V100 GPUs.

    \paragraph{Rebalanced dataset fine-tuning.} 
    We utilize the rebalanced dataset, \textit{FFHQ+LPFF-rebal}, to fine-tune $S^{Ours}_{var1}$, and denote the rebalanced model as $S^{Ours}_{var2}$.
    All training parameters are identical to those of $S^{Ours}_{var1}$.
    Our fine-tuning time is $\sim$18 hours on  8 Tesla V100 GPUs.

\subsection{EG3D}
\label{subsec:EG3D_retrain}

The mapping network, volume rendering module, and dual discriminator in EG3D~\cite{Chan_2022_CVPR} are all camera pose-dependent. We divided the EG3D model into three modules: Generator $G$, Renderer $R$, and Discriminator $D$, 
{please refer to the supplement file for an illustration of the three modules.}
The attribute correlations between pose and other semantic attributes in the dataset are faithfully modeled by using the camera parameters fed into $G$. $R$ and $D$ are always fed with the same camera specifications. The camera parameters help $D$ ensure multi-view-consistent super resolution and direct $R$ in how to render the final images from various camera views.

    In this paper, we define two types of camera parameters that are inputted into the whole model as:
    \begin{equation}
            c = [c_{g},c_{r}],
            \label{eq:camera_presentation}
        \end{equation} 
    where $c_{g}$ stands for the camera parameters fed into $G$, and $c_{r}$ stands for the camera parameters fed into $R$ and $D$. $c_g$ will influence the face geometry and appearance and should be fixed in testing. 
    The authors of EG3D discover that maintaining $c_{g}=c_{r}$ throughout training can result in a GAN that generates 2D billboards.
    To solve this problem, they apply a swapping strategy that randomly swaps 
    $c_g$ with another random pose in that dataset with $\beta$ probability, where $\beta$ is a hyper-parameter.
    
    
    \paragraph{Retrain.}
    We use \textit{FFHQ+LPFF} to train EG3D from scratch.
    All the training parameters are identical to those of EG3D,
    where $\beta$ is linearly decayed from 100\%to 50\% over the first 1M images, and then fixed as 50\% in the remaining training.
    We denote the EG3D trained on \textit{FFHQ} as $E^{FFHQ}_{var1}$ (the original EG3D), denote the EG3D trained on \textit{FFHQ+LPFF} as $E^{Ours}_{var1}$.
     Our training time is $\sim$6.5 days on  8 Tesla V100 GPUs. 

    \paragraph{Rebalanced dataset fine-tuning.}
     In EG3D, the authors use the rebalanced dataset \textit{FFHQ-rebal} to fine-tune $E^{FFHQ}_{var1}$, leading to a more balanced model. We denote the fine-tuned model as $E^{FFHQ}_{var2}$. 
     For a fair comparison, we also use the same fine-tuning strategy as EG3D to  fine-tune our model $E^{Ours}_{var1}$ on our rebalanced dataset \textit{FFHQ+LPFF-rebal}.
    %
    $\beta$ is fixed as 50\% in training, and other training parameters are identical to those of EG3D.
    %
    We denote $E^{Ours}_{var1}$ fine-tuned on \textit{FFHQ+LPFF-rebal} as $E^{Ours}_{var2}$.
    Our fine-tuning time is $\sim$1 day on  8 Tesla V100 GPUs. 


\section{Evaluation}
\label{sec:Evaluation}
{To show that \textit{LPFF} can help 2D and 3D-aware face generators generate realistic results across large poses, we will first evaluate the performance of 2D face generators (Sec.~\ref{sec: eval_stylegan}), and then demonstrate the performance of 3D-aware face generators (Sec. \ref{sec: eval_EG3D}).}
\subsection{StyleGAN}
\label{sec: eval_stylegan}
      \begin{figure}[t] 
          \centering
          \includegraphics[width=.85\columnwidth]{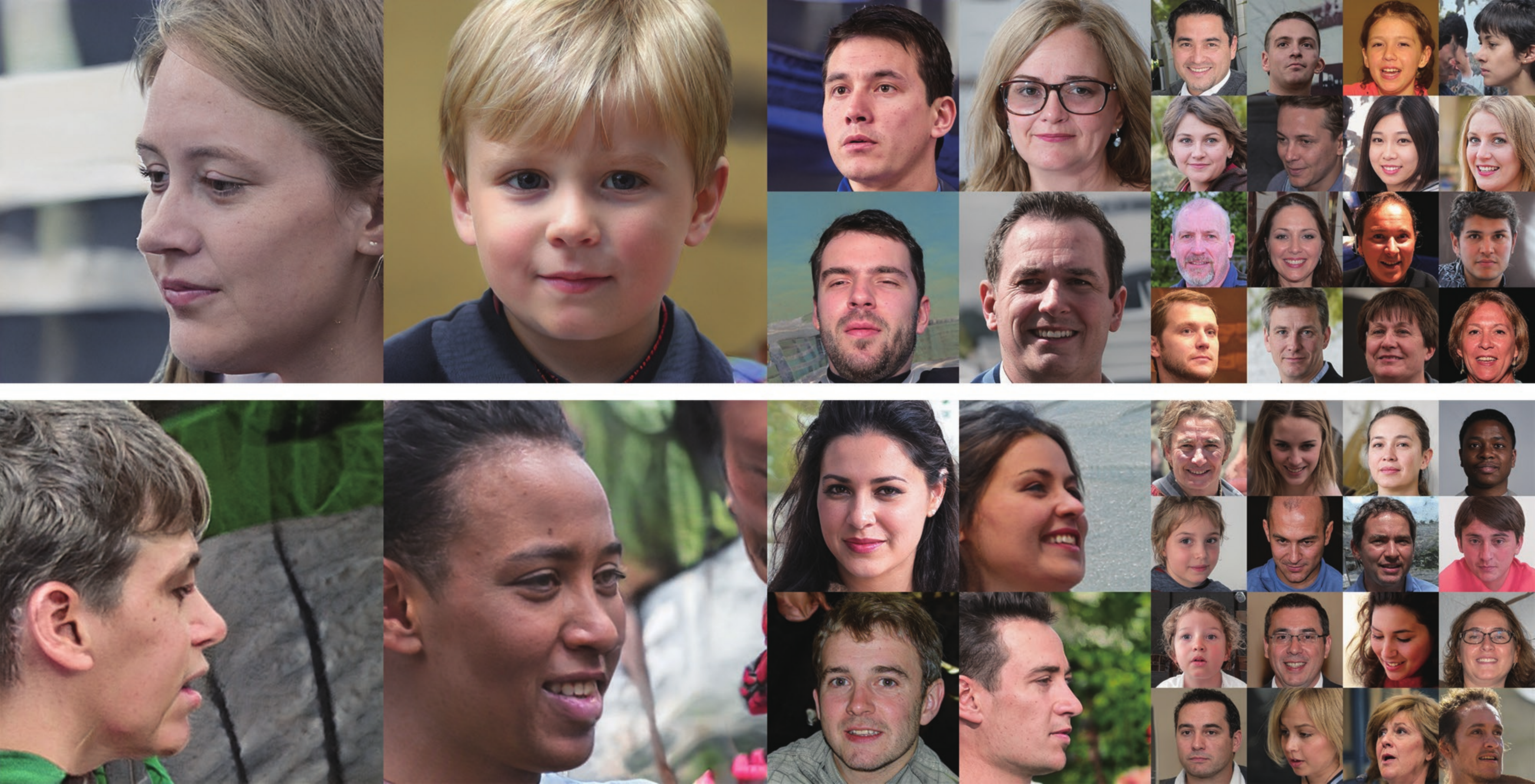}
          \caption{Images produced by our $S^{Ours}_{var1}$ model (Top) and $S^{Ours}_{var2}$ model (Bottom). We apply truncation  with $\psi=0.7$.}
          \vspace{-5pt}
          \label{fig:stylegan_gen}
        \end{figure}
    
    Fig.~\ref{fig:stylegan_gen} shows the uncurated samples of faces generated by the models trained on our dataset, with resolution $1024^2$. Our models synthesize images that are of high quality and have large pose variance.

    \paragraph{FID and perceptual path length.}  We trained the models using different datasets, so the latent space distributions are different in our experiments. Therefore, we do not compare 
    the Fr\'echet Inception Distance (FID) \cite{DBLP:conf/nips/HeuselRUNH17} and perceptual path length (PPL) \cite{DBLP:conf/cvpr/KarrasLA19} between 
    the models, since they are highly related to dataset distributions. 
    Instead, we respectively measure the FID of $S^{FFHQ}_{var1}$, $S^{Ours}_{var1}$ and  $S^{Ours}_{var2}$ on their training dataset. The FID of $S^{FFHQ}_{var1}$ is 2.71 on \textit{FFHQ}, the FID of $S^{Ours}_{var1}$ is 3.407 on \textit{FFHQ+LPFF}, and the FID of $S^{Ours}_{var2}$ is 3.786 on \textit{FFHQ+LPFF-rebal}. 
    {The comparable FIDs show that the StyleGAN2-ada model can achieve convergence on our datasets as it did on \textit{FFHQ}.}
    We use the PPL metric that is computed based on path endpoints in $W$ latent space, without the central crop.  The PPL of $S^{FFHQ}_{var1}$ is 144.9, the PPL of $S^{Ours}_{var1}$ is 147.6, and the PPL of $S^{Ours}_{var2}$ is 173.0.
    The PPL of $S^{Ours}_{var1}$ is comparable to the PPL of $S^{FFHQ}_{var1}$. The higher PPL of $S^{Ours}_{var2}$ indicates that $S^{Ours}_{var2}$ leads to more drastic image feature changes when performing interpolation in the latent space. We 
    attribute this to the larger pose variance in $S^{Ours}_{var2}$'s latent space and the \textit{FFHQ+LPFF-rebal} dataset.

  \begin{figure}[t]
    	\centering
    	{\includegraphics[width=0.95\columnwidth]{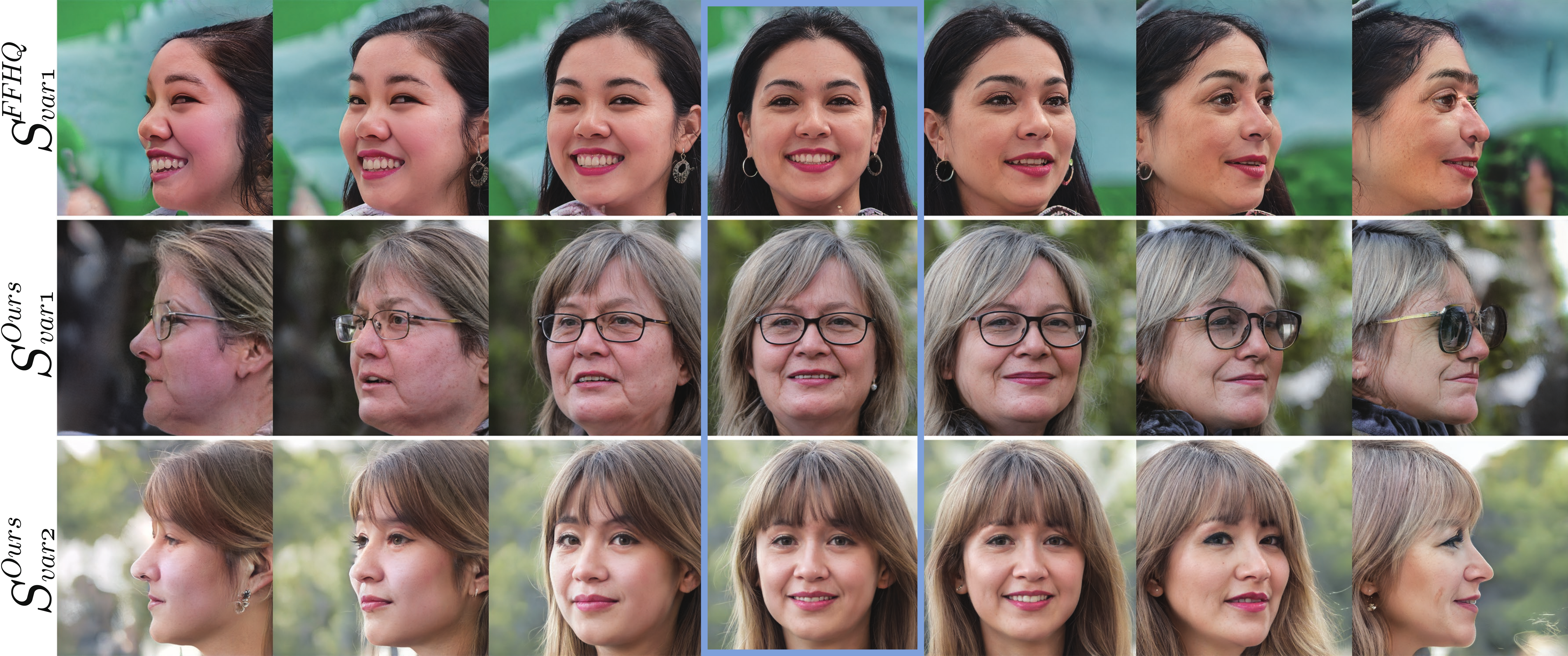}}
    	\caption{Pose manipulation comparison between $S^{FFHQ}_{var1}$ (Top), $S^{Ours}_{var1}$ (Middle), and $S^{Ours}_{var2}$ (Bottom). The images highlighted by the blue box are generated from randomly sampled latent codes, and all the samples are linearly moved along the yaw editing direction with the same distance.
    	}
    	\vspace{-5pt}
    	\label{fig:stylegan_interface_pose_comparison}
    \end{figure}
    

\paragraph{Pose manipulation.}
    We compare the pose distribution of the latent spaces by displaying the results of linear yaw pose manipulation. 
    {For each model, we label randomly sampled latent codes according to the camera parameters of the corresponding synthesized images (yaw angles $\textgreater 90^{\circ}$ as positive and $\leq 90^{\circ}$ as negative) and use InterfaceGAN \cite{DBLP:conf/cvpr/ShenGTZ20} to compute the yaw editing direction. 
    The pose editing results are then obtained by moving randomly sampled latent codes along the yaw editing direction, as shown in Fig. \ref{fig:stylegan_interface_pose_comparison}.
    Because the linear manipulation method is used without any semantic attribute disentanglement, the results of all models cannot preserve facial identity.}
    As for $S^{FFHQ}_{var1}$, the ``side face'' results are far from a genuine human face, demonstrating that the latent codes have reached the edge of the latent space.
    With regard to  $S^{Ours}_{var1}$ and $S^{Ours}_{var2}$, our models produce reasonable and {comparable} large-pose portraits. 
    The comparison shows that our models' latent spaces are more extensive and better able to represent large-pose data.

\begin{figure}[t]
    	\centering
    	{\includegraphics[width=.8\columnwidth]{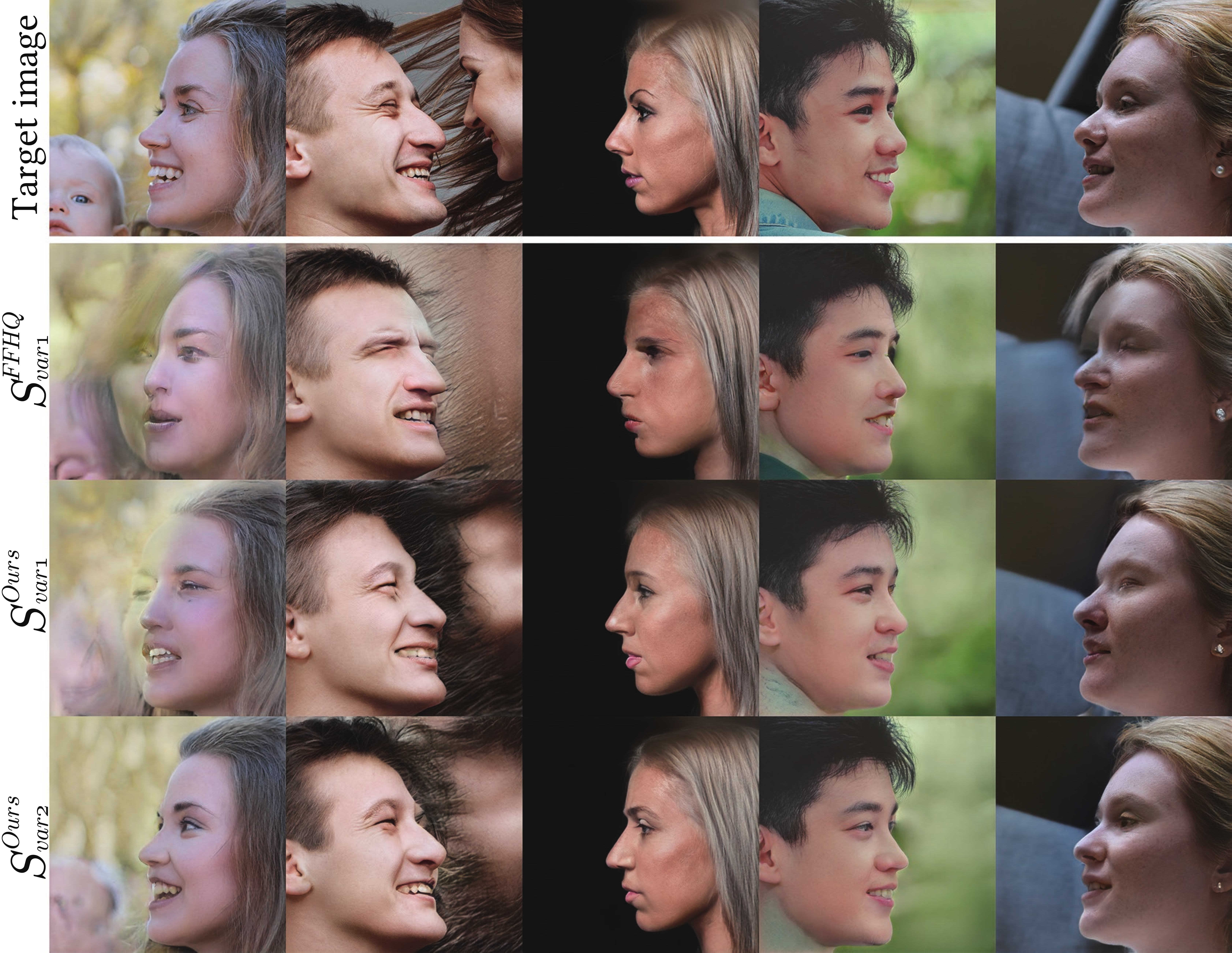}}
    	\caption{Large-pose data projection comparison between $S^{FFHQ}_{var1}$, $S^{Ours}_{var1}$, and $S^{Ours}_{var2}$. The target images (the first row) are collected from Unsplash and Pexels websites.}
    	\vspace{-5pt}
    	\label{fig:stylegan_avg_init_projection}
    \end{figure}

 \paragraph{Large-pose data inversion and manipulation.}
    To further show that our models can better represent large-pose data, we project large-pose portraits into the latent spaces of those models (see Fig.~\ref{fig:stylegan_avg_init_projection}), and apply semantic editing to the obtained latent codes.
    We collect the testing images from Unsplash\footnote{\hyperref[]{https://unsplash.com}} and Pexels\footnote{\hyperref[]{https://www.pexels.com}} (independent of both \textit{FFHQ} and \textit{LPFF}). We then employ 500-step latent code optimization in $W+$ latent space to minimize the distance between the synthesized image and the target image. 
    To evaluate the editability of the projected latent codes, we use the attribute classifiers provided by StyleGAN \cite{DBLP:conf/cvpr/KarrasLA19} and employ InterfaceGAN to compute semantic boundaries for each model, and then use the boundaries to edit the projected latent codes.
    We also use the yaw editing directions to try to make the large pose data face forward.
    Please refer to the supplement for those semantic editing results. 
    As shown in those projection and manipulation results, the models trained on our dataset have fewer artifacts and can better represent the large pose data in their latent spaces. 
    {What's more, $S^{Ours}_{var2}$ outperforms $S^{Ours}_{var1}$ because $S^{Ours}_{var2}$ is trained on a more balanced dataset, which proves the effectiveness of our data rebalance strategy.
    }
    
      \begin{figure*}

\begin{minipage}[b]{.65\linewidth}
    
    \includegraphics[width=1.0\linewidth]{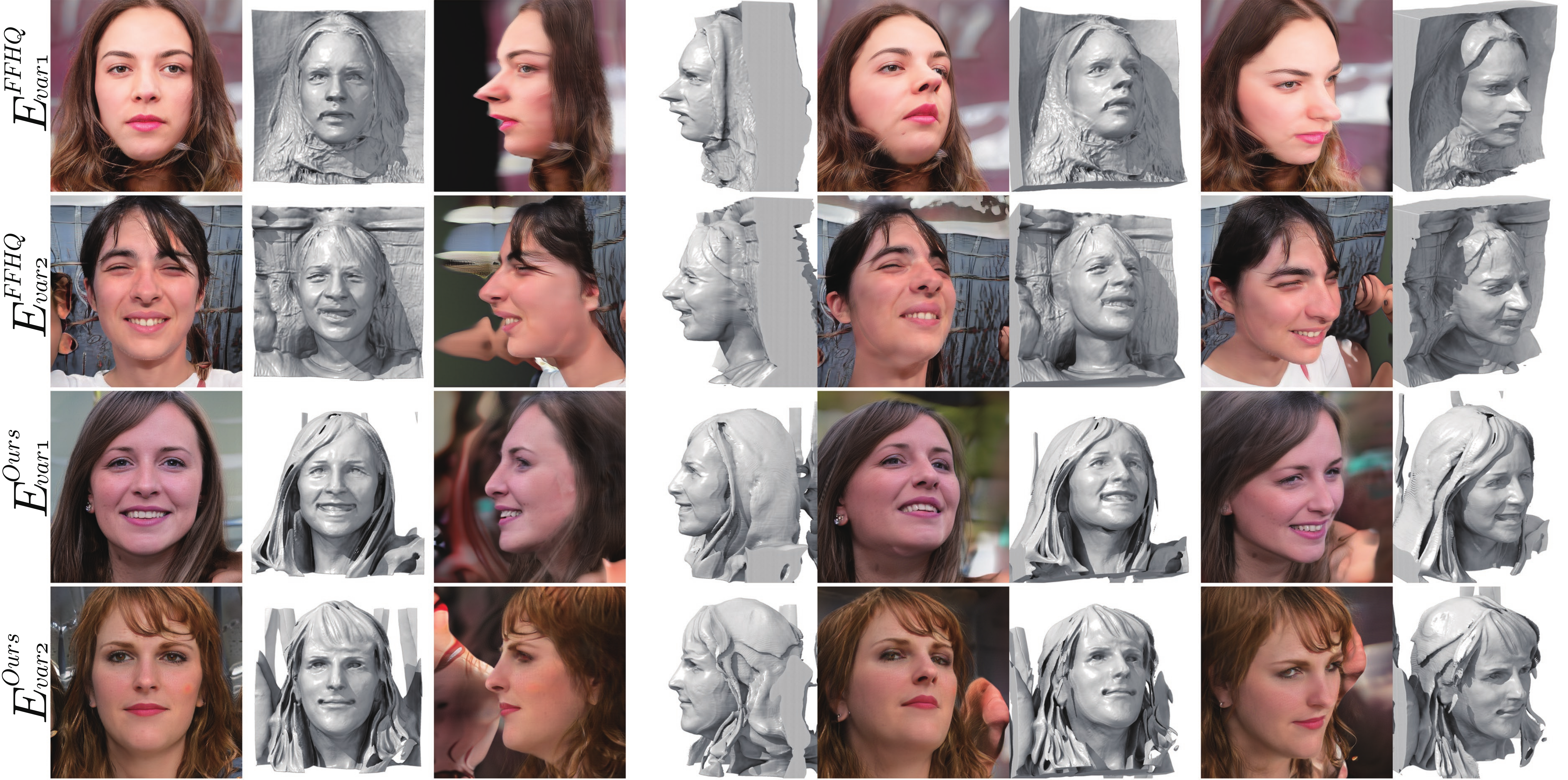}
          \caption{
          {Image-shape pairs produced by $E^{FFHQ}_{var1}$,$E^{FFHQ}_{var2}$, $E^{Ours}_{var1}$, and $E^{Ours}_{var2}$.
          We apply truncation  with $\psi=0.8$.}
          }
          \label{fig:eg3d_gen_curated}
\end{minipage}
\medskip
 \hfill
 \begin{minipage}[b]{.32\linewidth}
    \centering
   \scalebox{0.7}{
    \begin{tabular}{@{}cccc@{}}
    \toprule
    \multicolumn{1}{l}{\multirow{2}{*}{model}} & \multicolumn{1}{l}{$c_g =c_{avg}$} & \multicolumn{1}{l}{$c_g \sim$ FFHQ}  & \multicolumn{1}{l}{$c_g \sim$ LPFF}  \\
    \multicolumn{1}{l}{}   & $c_r \sim$ FFHQ     & $c_r \sim$FFHQ & $c_r \sim$FFHQ    \\ \midrule
    $E^{FFHQ}_{var1}$ & 0.771  &	0.768 &	0.760   	 \\
    $E^{Ours}_{var1}$ &\textbf{0.804}&\textbf{0.792}&\textbf{0.778} \\ \midrule
    $E^{FFHQ}_{var2}$ &0.770&0.769&0.766   \\
    $E^{Ours}_{var2}$ &\textbf{0.789}&\textbf{0.784}&\textbf{0.771}\\ \bottomrule
    \end{tabular}
    }
            \captionof{table}{ 
            {Quantitative evaluation of facial identity consistency ($\uparrow$). }
            }
        \label{tab:Facial_identity}
 \vspace{5pt}
    \scalebox{0.7}{
        \begin{tabular}{@{}cccc@{}}
        \toprule
        \multicolumn{1}{l}{\multirow{2}{*}{model}} & \multicolumn{1}{l}{$c_g =c_{avg}$} & \multicolumn{1}{l}{$c_g \sim$ FFHQ}  & \multicolumn{1}{l}{$c_g \sim$ LPFF}  \\
        \multicolumn{1}{l}{}   & $c_r \sim$ FFHQ     & $c_r \sim$FFHQ & $c_r \sim$FFHQ    \\ \midrule
        $E^{FFHQ}_{var1}$ & 0.134&0.133&0.159	 	 \\
        $E^{Ours}_{var1}$ &\textbf{0.119}&\textbf{0.124}&\textbf{0.134} \\ \midrule
        $E^{FFHQ}_{var2}$ &0.135&0.130&0.142    \\
        $E^{Ours}_{var2}$ &\textbf{0.117}&\textbf{0.122}&\textbf{0.131}  \\  \bottomrule
        \end{tabular}
        }
                \captionof{table}{{
                    Quantitative evaluation of geometry consistency ($\downarrow$).
                }
                }
            \label{tab:Geometry}
\end{minipage} 

\end{figure*}

\begin{table*}[t]
\centering
\scalebox{0.7}{
    \begin{tabular}{@{}ccccccccc@{}}
    \toprule
    \multicolumn{1}{l}{\multirow{2}{*}{model}} & \multicolumn{1}{l}{$c_g =c_{avg}$} & \multicolumn{1}{l}{$c_g =c_{avg}$} & \multicolumn{1}{l}{$c_g \sim$ FFHQ} & \multicolumn{1}{l}{$c_g \sim$ FFHQ} & \multicolumn{1}{l}{$c_g \sim$ LPFF} & \multicolumn{1}{l}{$c_g \sim$ LPFF} & \multicolumn{1}{l}{$c_g \sim$ FFHQ} & \multicolumn{1}{l}{$c_g \sim$ LPFF}\\
    \multicolumn{1}{l}{}   & $c_r \sim$ FFHQ     & $c_r \sim$ LPFF  & $c_r \sim$FFHQ & $c_r \sim$LPFF   & $c_r \sim$FFHQ & $c_r \sim$LPFF & $c_r =c_g$ &  $c_r =c_g$    \\ \midrule
    $E^{FFHQ}_{var1}$ &\textbf{6.523}&23.598&\textbf{4.273}&22.318&23.698&36.641&\textbf{4.025}&23.301   \\
    $E^{Ours}_{var1}$ & 7.997&\textbf{20.896}&6.623&\textbf{19.738}&\textbf{21.300}&\textbf{22.074}&6.093&\textbf{16.026}  \\ \midrule
    $E^{FFHQ}_{var2}$ &\textbf{6.589}&20.081&\textbf{4.456}&19.983&19.469&30.181&\textbf{4.262}&23.717   \\
    $E^{Ours}_{var2}$ & 9.829&\textbf{16.775}&6.672&\textbf{15.047}&\textbf{13.022}&\textbf{14.836}&6.571&\textbf{12.221} \\ \bottomrule
    \end{tabular}
    }
            \caption{FID ($\downarrow$) for EG3D generators that are trained on different datasets. We calculate the FIDs by sampling 50,000 images using different sampling strategies and different camera distributions. 
            {We compare the models that are trained with the same training strategy ($var1/var2$).
            }
            \vspace{-5pt}
            }
        \label{tab:FID1}
\end{table*}

    \subsection{EG3D}   
    \label{sec: eval_EG3D}
      Fig.~\ref{fig:eg3d_gen_curated} provides the selected samples that are generated by the models trained {on the \textit{FFHQ} dataset and our dataset}, with resolution $512^2$. Even in large poses, our synthesized images and 3D geometry are high-quality.
      
              \paragraph{FID.}
        \label{subsec:FID}
        In EG3D, the generator is conditioned on a fixed camera pose ($c_{g}$) when rendering from a moving camera trajectory to prevent the scene from changing when the camera ($c_r$) moves during inference. However, EG3D's authors evaluated the FID of EG3D by conditioning the model on $c_{g}$ and rendering results from $c_r = c_g$. This approach cannot demonstrate the performance of multi-view rendering during inference, since
        the generator always ``sees'' the true pose of the rendering camera in evaluation, but omits other poses. 
        For a 3D-aware generator, we are more interested in how a face looks from various camera views (which can indicate the quality of face geometry to some extent).
        So a more reasonable way is to let $c_r$ and $c_g$ be independent of
        each other and sample them from the respective  distributions that are of our interest.
        To achieve this, we propose a novel FID measure, which is based on
        three camera sampling strategies. First, we fix 
        $c_{g}$ as $c_{avg}$ and then sample $c_r$ from different datasets. Second, we respectively sample $c_{r}$ and $c_{g}$ from different datasets. Third, we sample $c_{g}$ from different datasets and set $c_r=c_g$ (the one that was used in EG3D). See the calculated FID values in Tab. \ref{tab:FID1}.

        Models trained on our datasets exhibit improvements in FID in most cases, particularly when the final results are rendered from large poses ($c_r \sim LPFF$), or when the generator is conditioned on large poses ($c_g \sim LPFF$).  
        {We notice that there is an increased FID when computing $c_g=c_{avg}/c_r, c_r\sim FFHQ$.}
        As explained by the authors of EG3D,  the pre-trained $E^{FFHQ}_{var1}$ and $E^{FFHQ}_{var2}$ were achieved using buggy (XY, XZ, ZX) planes.  
        In our experiments, since we fix this bug as they suggested using (XY, XZ, ZY),
        the XZ-plane representation's dimension would be cut in half, thus weakening the expressive capacity for frontal faces.

        Thanks to our dataset rebalancing strategy, $E^{Ours}_{var2}$ can pay more attention to large pose data and enhance the rendering quality, thus further improving the FID of $E^{Ours}_{var1}$ on large poses. 
        {When computing FID of $c_g=c_{avg},c_r\sim FFHQ$, we notice that $E^{Ours}_{var2}$ has an increased FID compared to $E^{Ours}_{var1}$, while $E^{FFHQ}_{var2}$ and  $E^{FFHQ}_{var1}$ have comparable results. 
        This is due to the addition of new large-pose data, \textit{LPFF}.
        FID is highly related to the data distribution, and the rebalancd \textit{FFHQ+LPFF-rebal} dataset changes the data distribution when rendering from medium poses.
        }

    

          \begin{figure*}[h]
            \centering
            \includegraphics[width=0.85\textwidth]{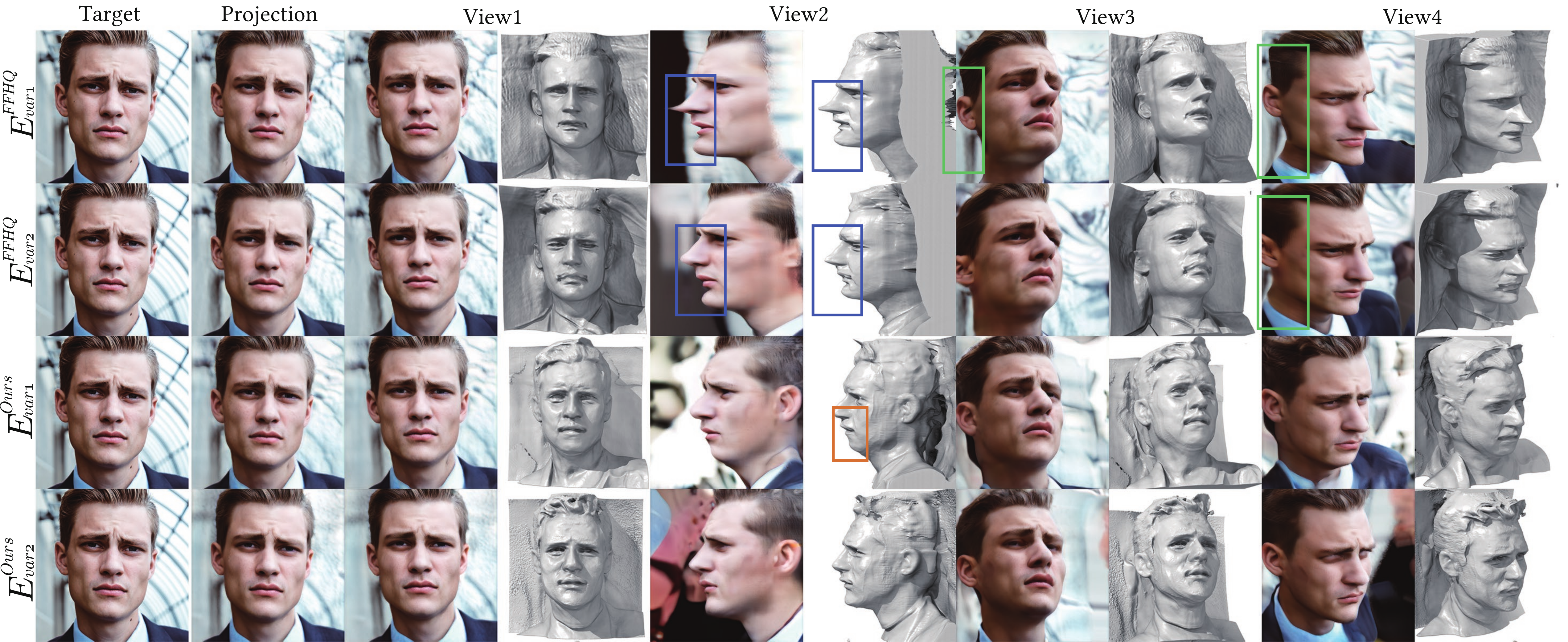}
                  \caption{
                To fit the single-view testing image, we employ HFGI3D \cite{xie2022high}. The obtained latent codes are then rendered using four novel views. The inversion is carried out in $W$ space, and the generators are conditioned on $c_{avg}$.
                   }
                  \label{fig:Unsplash_Pexels_4}
        \end{figure*}

        \paragraph{Facial identity consistency.}
         \label{subsec:Facial_Identity}   
        We leverage ArcFace \cite{DBLP:conf/cvpr/DengGXZ19} to measure the models' performance on facial identity maintenance. We render two novel views for 1,024 random faces and use ArcFace to compute the mean identity similarity for all image pairs.
        %
        We employ three sampling strategies for $c_g$ to evaluate the generator's performance on the camera distribution of different datasets. As for $c_r$, we find that the extreme rendering camera views will heavily influence the performance of ArcFace, so we only sample $c_r$ from the \textit{FFHQ} dataset, where most of the faces have small to medium poses. As shown in Tab. \ref{tab:Facial_identity}, our models present significant improvements in facial identity consistency across different sample strategies and datasets.

          \paragraph{Geometry consistency.}
        \label{subsec:Geometry}   
        We employ $\mathcal{F}$, which outputs 3DMM coefficients to evaluate the geometry consistency.
        We employ the same camera sampling methods as in facial identity consistency computation.
        We first render two novel views for 1,024 random faces. Then
        for each image pair, we compute the mean L2 distance of the face id and expression coefficient.
        As shown in Tab. \ref{tab:Geometry}, our models present improvement in geometry consistency across different sample strategies and datasets.

         \paragraph{Image inversion.}
         To evaluate the ability to fit multi-view images, we use FaceScape \cite{DBLP:conf/cvpr/Yang0WHSYC20} as the testing data.
         We use four multi-view images (including one with a small pose) of a single identity as the reference images. We perform latent code optimization to simultaneously project one or four images into $W+$ latent space. Then we use the camera parameters that are extracted from another 4 multi-view images to render novel views. Please refer to the supplement for multi-view image inversion results.
         {
         Because occluded face parts are unavoidable in single-view portraits, we perform single-view image inversion using HFGI3D \cite{xie2022high}, a novel method that combines pseudo-multi-view estimation with visibility analysis.
         }
         As shown in Fig.~\ref{fig:Unsplash_Pexels_4}, the inversion results indicate that $E^{FFHQ}_{var1}$ and $E^{FFHQ}_{var2}$  suffer from the ``wall-mounted'' unrealistic geometry.
         Due to the adhesion between the head and the background, there are missing ears in View 2 and distorted ears and necks in Views 3 and 4 (highlighted by green boxes). A pointed nose exists in View 2 (highlighted by blue boxes).
         Our $E^{Ours}_{var1}$ and $E^{Ours}_{var2}$ models produce reconstructed face geometry that is free from those artifacts, suggesting that the learned 3D prior from our dataset is more realistic. It also shows that after employing the data rebalance in Sec. \ref{subsec:Data Rebalance}, lips are more natural in $E^{Ours}_{var2}$ compared to $E^{Ours}_{var1}$ (highlighted by an orange box).


        \paragraph{``Seam'' artifacts.}
        The authors of IDE-3D speculate that the ``seam"  artifacts in EG3D could be caused by the imbalanced camera pose distribution of datasets, and propose a density regularization loss to deal with the ``seam"  artifacts along the edge of the faces. 
        Compared to the IDE-3D, our model $E^{Ours}_{var1}$ is trained without requiring any additional regularization loss or any data rebalance strategy, and is free from the ``seam" artifacts.  Please refer to the supplement for the illustration of ``seam" artifacts.

\section{Conclusion}
In order to address the pose imbalance in the current face generator training datasets, we have presented \textit{LPFF}, a large-pose Flickr face dataset comprised of 19,590 high-quality real large-pose portrait images. 
Compared to those models trained on \textit{FFHQ}, the 2D face generators trained on our dataset display a latent space that is more representative of large poses and achieve better performance when projecting and manipulating large-pose data. The 3D-aware face generators trained on our dataset can produce more realistic face geometry and render higher-quality results at large poses. The rendering results are also more view-consistent.
We hope our dataset can inspire more portrait generating and editing works in the future. 

Our work has several limitations.
Despite having a more balanced camera pose distribution, our dataset still has a semantic attribute imbalance. For instance, we measured the probability of smiling in \textit{FFHQ+LPFF}. The plot shows that people typically smile when they are facing the camera, so the models trained on our dataset have the smile-posture entanglement. 
{Please refer to the supplement file for the smiling probability plot.}
{This can be overcome by building a camera system and capturing large-scale semantic-balanced images.}
{
Our processing pipeline uses the face detector and face reconstruction model to align faces, but it does not perform well under extreme conditions (for example, when only the back of the head is visible and the face is completely occluded). As a result, we cannot obtain full-head results.}
Future work that can model the full head may be helpful to get a more extensive dataset.

{\small
\bibliographystyle{ieee_fullname}
\bibliography{egbib}
}

\end{document}